\documentclass{article}

\usepackage[preprint]{neurips_2026}

\usepackage[utf8]{inputenc} 
\usepackage[T1]{fontenc}    
\usepackage{hyperref}       
\usepackage{url}            
\usepackage{booktabs}       
\usepackage{amsfonts}       
\usepackage{nicefrac}       
\usepackage{microtype}      
\usepackage{xcolor}         
\usepackage{graphicx}
\usepackage{amsmath}
\usepackage{amssymb}
\usepackage{bm}
\usepackage{wrapfig}
\usepackage{makecell}
\usepackage[ruled,vlined]{algorithm2e}
\newcommand{\method}{PhySPRING}
\title{\method{}: Structure-Preserving Reduction of Physics-Informed Twins via GNN}

\author{%
  Yixiong Jing\thanks{Equal contribution.} \\
  University of Cambridge\\
  \texttt{yj401@cam.ac.uk}
  \And
  Xingyuan Chen\footnotemark[1] \\
  Shanghai Jiao Tong University\\
  \texttt{February25th@sjtu.edu.cn}
  \And
  Guangming Wang\thanks{Corresponding author.} \\
  University of Cambridge\\
  \texttt{gw462@cam.ac.uk}
  \AND
  Olaf Wysocki \\
  University of Cambridge\\
  \texttt{okw24@cam.ac.uk}
  \And
  Haibing Wu \\
  University of Cambridge\\
  \texttt{hw657@cam.ac.uk}
  \And
  Brian Sheil \\
  University of Cambridge\\
  \texttt{bbs24@cam.ac.uk}
}

\begin{document}

\maketitle


\begin{abstract}
Physics-based digital twins aim to predict the dynamics of real-world objects under interaction, enabling real-to-sim-to-real applications in robotics. Current approaches reconstruct such twins as explicit physical models (such as spring-mass systems) to predict the dynamics, but the resulting models often inherit the resolution of the visual reconstruction rather than being reduced to the physical complexity required to reproduce task-relevant dynamics. This mismatch introduces redundant topology, making repeated forward-dynamics rollouts unnecessarily expensive.
To address this challenge, we present \method{}, an fully differentiable GNN-based method to reduce complexity in spring--mass digital twins. \method{} jointly learns a hierarchy of coarsened graph topologies and their mechanical parameters from observations. At each reduction level, \method{} merges nodes with similar learned dynamic responses to optimize the topology, while maintaining every reduced layer as an explicit spring--mass system. 
On the PhysTwin benchmark, \method{} improves dense reconstruction and prediction accuracy over PhysTwin, while reduced models retain stable physical and visual fidelity with up to a 2.30$\times$ speed-up. 
We further demonstrate the effectiveness of \method{} in a Real2Sim robot policy-evaluation pipeline, where the reduced models are substituted zero-shot into ACT and $\pi_0$ evaluations, maintaining comparable manipulation success rates across downsampling levels while improving action-sampling effectiveness. Together, \method{} enables efficient and structure-preserving spring--mass reduction without sacrificing fidelity or robotic utility.

\end{abstract}

\section{Introduction}

Reconstructing physics-informed digital twins from real-world objects enables accurate prediction of how objects move and deform under interaction~\cite{xian2025empm, jiang2025phystwin}, supporting downstream applications such as robot manipulation planning~\cite{2026arXiv260301151L,
2025arXiv251008568L, pmlr-v270-li25c, jangir2025robotarena, guo2025ctrl} and data generation for policy learning~\cite{chebotar2019closing, muratore2022neural, 2025arXiv251104665Z}. However, precise physical reconstruction from observations remains non-trivial. Previous graphics and engineering pipelines rely on carefully designed constitutive models and discretizations~\cite{sifakis2012fem, weiss2023fast} that assume comprehensive prior knowledge of object physics, making them expensive and difficult to generalize.

Leveraging recent advances in differentiable physics~\cite{Macklin_Warp_A_High-performance_2022} and 3D reconstruction~\cite{kerbl20233d}, PhysTwin~\cite{jiang2025phystwin} learns a
spring--mass physical model directly from RGB-D video via differentiable
simulation, with appearance reconstructed by 3D Gaussian Splatting (3DGS)~\cite{kerbl20233d}.
However, current physics-driven reconstruction approaches directly inherit physical simulation resolution from visual reconstruction resolution. The resolution needed to represent surface geometry and appearance can be much higher than the number of degrees of freedom needed to reproduce task-relevant dynamics for downstream robotics applications. Planning and policy evaluation require repeated forward-dynamics rollouts under varied interactions, so simulating a dense physical representation inside the robot loop can introduce unnecessary computational overhead. Therefore, we identify that the core challenge is to obtain a reduced-order physical representation that accelerates simulation while preserving physical fidelity and visual/geometric consistency.

One possible approach is to learn a reduced topology with GNN-based simulation models~\cite{pfaff2020learning, evomesh_icml2024}. These methods relax discrete topology-reduction decisions into end-to-end trainable architectures, enabling adaptive graph representations that reduce physical-system complexity for dynamics prediction. However, they typically require ground-truth dynamic states from numerical solvers and learn black-box forward dynamics rather than explicit
physical parameters. As a result, the learned reduced models do not preserve the same physical structure as the original system, and therefore cannot be directly deployed in external simulators under unseen interactions such as actions from robot arms.

Classical model order reduction (MOR) provides a physics-based approach for reducing physical systems through nodal clustering based on analytically computed dynamic similarity~\cite{model_reduction_2017, antoulas2005approximation}. This preserves the second-order mechanical form of the system and produces reduced graphs that can be integrated by standard numerical solvers. However, these methods assume known physical matrices and material parameters, which requires pre-optimization with observation data. Moreover, computing dynamic-similarity matrices over thousands of reconstructed nodes is computationally expensive, limiting their practicality for downstream applications.

\begin{figure}[t!]
\centering
\includegraphics[width=\textwidth]{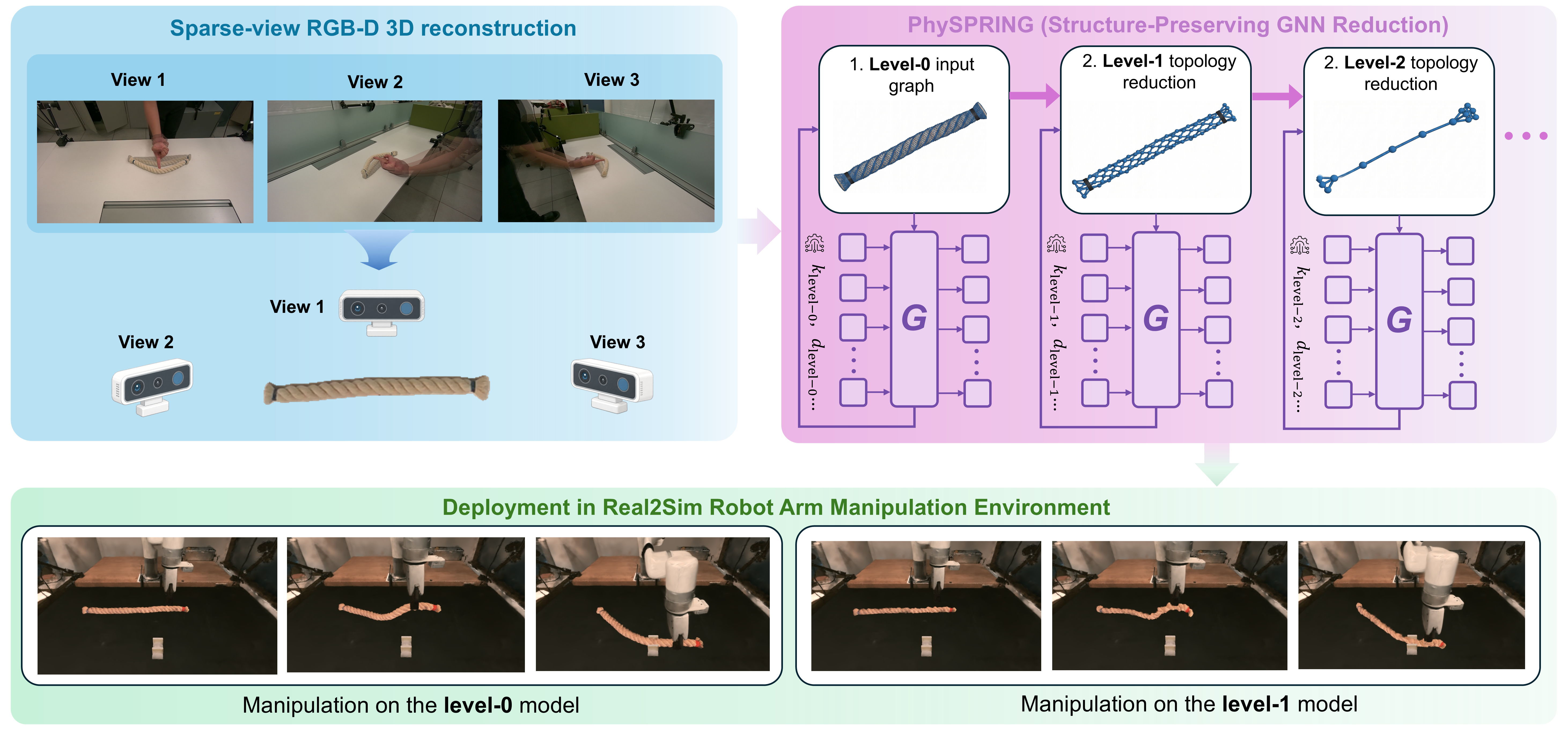}
\caption{We propose \method{}, a novel method for structure-preserving order reduction of physics-informed digital twins reconstructed from sparse-view RGB-D observations. An end-to-end multi-hierarchical GNN jointly predicts the reduced topology and the per-edge mechanical parameters at every level $l$. Since each reduced-order model remains an explicit second-order spring--mass system, it is directly deployable in a downstream Real2Sim robot-arm manipulation simulator~\cite{2025arXiv251104665Z}.}
\label{fig:overview}
\end{figure}

To address this gap, we present \textbf{\method{}} a structure-preserving reduction method for
spring--mass digital twins reconstructed from sparse-view RGB-D observations as shown in Fig.~\ref{fig:overview}. \method{} combines GNN-based topology reduction with differentiable physics so
that the reduced model remains an explicit physical system rather than a black-box dynamics predictor. Our contributions are as follows.
\textbf{(i)}~We introduce an end-to-end physics-informed GNN that jointly learns topology reduction and mechanical parameters from observation-supervised rollouts. On PhysTwin, \method{} improves dense reconstruction/prediction over PhysTwin, and reduced models achieve controlled fidelity--efficiency trade-offs with up to 2.30$\times$ speed-up.
\textbf{(ii)}~We preserve the explicit spring--mass form of the digital twin
across all reduction levels. Galerkin projection induces coarse spring-mass representation while a GNN decoder refines the resulting mechanical parameters through differentiable rollouts.
\textbf{(iii)}~We evaluate \method{} in a Real2Sim~\cite{2025arXiv251104665Z} policy-evaluation pipeline. The reduced models can be substituted zero-shot into ACT~\cite{zhao2023learning} and $\pi_0$~\cite{black2024pi_0} without retraining, preserving manipulation success rate, while improving action-sampling throughput by up to 1.23$\times$ over the original model.


\section{Related works}

\paragraph{Physics-driven 4D Reconstruction} 
Recent 4D reconstruction methods achieve high-fidelity geometry and appearance from visual observations~\cite{kerbl20233d,mildenhall2021nerf,pumarola2021dnerf,park2021nerfies,wu20244dgs,yang2024gaussianflow}. However, they are primarily descriptive and lack physical priors for predicting future dynamics or supporting grounded interaction. To improve physical fidelity, several works combine reconstructed scenes with explicit numerical simulation~\cite{xie2024physgaussian,zhang2024physdreamer,feng2024pie,xie2025physanimator} or infer material properties from video via differentiable physics~\cite{li2023pacnerf,gao2025seeing}. These methods often rely on predefined materials or implicit geometry and tend to prioritize visual plausibility over accurately recovering the dynamic behavior of real-world objects. To jointly recover both 3D structure and dynamics, other methods~\cite{zhong2024reconstruction,jiang2025phystwin,zhang2024dynamics} reconstruct physically grounded spring-mass models from observations to create physics-based digital twins. Nevertheless, these models inherit the geometric resolution of the observations, yielding redundant and over-constrained systems whose simulation cost limits their use inside robot planning loops.

\paragraph{Model order reduction} 
Classical model order reduction (MOR) compresses a dynamical system with known mechanical properties into a lower-order surrogate whose dynamic response closely approximates that of the original system~\cite{moore1981balanced, gugercin2008h_2, bai2002krylov, sirovich1987turbulence, willcox2002balanced, benner2015survey}. For second-order linear time-invariant systems, second-order Gramians reduce system complexity while preserving the dynamic structure (e.g., same form of governing equations)~\cite{chahlaoui2006second, reisstykel2008second}; a previous method~\cite{model_reduction_2017} adapts the method on spring-mass systems to conduct MOR via nodal clustering based on controllability-Gramian similarity, yielding reduced graphs compatible with a standard numerical integrator.
However, these methods require the mechanical properties of the system to measure nodal dynamic similarity, which cannot be directly acquired from observation. The resulting topology, once optimized, remains static throughout the dynamic sequence and is not updated temporally. Moreover, computing the dynamic-similarity matrices introduce expensive matrix operations, which become prohibitively costly as the system size grows.

\paragraph{Graph topology optimization for physical simulation}
NN-based simulators~\citep{sanchez2020gns,pfaff2020learning,li2019dpinet} learn structural dynamics from ground-truth solver rollouts, offering substantial speedups over traditional numerical solvers. While hierarchical extensions~\citep{lino2022multi,cao2023bsms,janny2023eagle,yu2024hcmt} further improve efficiency by pre-computing multi-scale coarse graphs, their heuristic topologies are fixed and cannot be optimized during the training. Differentiable graph pooling~\citep{ying2018diffpool,gao2019topkpool,lee2019sagpool} learns hierarchical graph representations via soft assignments or score-based node selection, yet is mainly designed for static graph-level prediction and does not enforce physical constraints, limiting fidelity in coarse dynamical models. EvoMesh~\citep{evomesh_icml2024} partly addresses this gap by using Gumbel--Softmax relaxation to jointly optimize topology and dynamics over time. However, it only predicts the forward dynamics without recovering mechanical parameters simultaneously. EvoMesh also produces topologies that are not physically consistent, limiting deployment in external simulators under unseen conditions.

\section{Methodology}

\subsection{Problem formulation}
\label{sec:problem_formulation}

Given multi-view RGB-D videos of a deformable object experiencing a mechanical interaction, our objective is to recover a hierarchy of physically
consistent digital twins, each governed by the same second-order
spring--mass dynamics but expressed on a progressively coarser graph.
Let $l\in\{0,1,\dots,L\}$ represent indices of the reduction level, with $l=0$ the
full-order model (FOM) at the geometric resolution of the observation
and $l=L$ the coarsest reduced-order model (ROM).

At level $l$, we discretize the object as a graph
$\mathcal{G}^{(l)} = \{\mathcal{V}^{(l)}, \mathcal{E}^{(l)}\}$ of $N_l = |\mathcal{V}^{(l)}|$ mass nodes connected by axial springs $\mathcal{E}^{(l)}$. The nodal positions and velocities of $\mathcal{G}^{(l)}$ at time $t$ are represented as $\mathbf{X}_t^{(l)},\,\dot{\mathbf{X}}_t^{(l)}\in\mathbb{R}^{N_l \times 3}$ and the comprehensive dynamic state is represented as
$\mathbf{Z}_t^{(l)} = (\mathbf{X}_t^{(l)},\,\dot{\mathbf{X}}_t^{(l)})$. The dynamics of $\mathcal{G}^{(l)}$ then follow the canonical second-order network system:
\begin{equation}
    \mathbf{M}^{(l)}\,\ddot{\mathbf{X}}_t^{(l)}
    \;+\;\mathbf{D}^{(l)}\,\dot{\mathbf{X}}_t^{(l)}
    \;+\;\mathbf{L}^{(l)}\,\mathbf{X}_t^{(l)}
    \;=\;
    \mathbf{f}_c^{(l)}\!\big(\mathbf{Z}_t^{(l)};\,\boldsymbol{\eta}^{(l)}\big)
    \;+\;\mathbf{u}_t^{(l)},
    \label{eq:second_order}
\end{equation}
where $\mathbf{M}^{(l)}$ is the diagonal mass matrix, $\mathbf{L}^{(l)}$ is the weighted graph Laplacian whose off-diagonal entries are the negated axial spring stiffnesses on $\mathcal{E}^{(l)}$~\citep{model_reduction_2017}, $\mathbf{D}^{(l)}$ is the damping matrix, $\mathbf{f}_c^{(l)}$ is the state-dependent contact force parameterized by learnable coefficients $\boldsymbol{\eta}^{(l)}$ (e.g., restitution and Coulomb friction), and $\mathbf{u}_t^{(l)}$ is the external manipulator input, restricted to the controller nodes in $\mathcal{V}^{(l)}$ and known from observation. The mass matrix $\mathbf{M}^{(0)}$ is fixed under the uniform-density assumption of PhysTwin~\citep{jiang2025phystwin} and is therefore not considered in the optimization. The remaining learnable parameters at level $l$ are represented in the tuple $\alpha^{(l)} \;=\; \big(\mathbf{L}^{(l)},\,\mathbf{D}^{(l)},\,\boldsymbol{\eta}^{(l)}\big)$, and the dynamics of $\mathcal{G}^{(l)}$ of Eq.~\eqref{eq:second_order} can be predicted via an Euler integrator $F^{(l)}$ using $\mathbf{Z}_{t+1}^{(l)} = F^{(l)}\!\big(\mathbf{Z}_t^{(l)},\,\mathbf{u}_t^{(l)}\big)$ with the level-$l$ ROM defined by $(\mathbf{M}^{(l)},\,\alpha^{(l)},\,\mathcal{G}^{(l)})$.

Our objective is to make the topology optimization of $\mathcal{G}^{(0)}$ differentiable by learning an assignment matrix $\mathbf{P}^{(l)}\in\{0,1\}^{N_l\times N_{l+1}}$ via a GNN. The coarser system matrices $(\mathbf{M}^{(l+1)},\,\mathbf{L}^{(l+1)},\,\mathbf{D}^{(l+1)})$ are then induced from $\mathbf{P}^{(l)}$ via Galerkin projection.

Let $\mathbf{X}_t^{\text{obs}}$ denote the partial 3D observation lifted from the RGB-D frames at time $t$. Our objective is to jointly estimate $\{\alpha^{(l)},\,\mathbf{P}^{(l)}\}_{l=0}^{L}$ by minimizing the discrepancy between simulated and observed dynamics at every level:
\begin{equation}
\resizebox{0.92\linewidth}{!}{$
\displaystyle
\min_{\{\alpha^{(l)},\,\mathbf{P}^{(l)}\}}\;
\sum_{l=0}^{L}\sum_{t=0}^{T}
\Big(
\mathcal{L}_{\text{geo}}(\hat{\mathbf{X}}_t^{(l)},\,\mathbf{X}_t^{\text{obs}})
+
\mathcal{L}_{\text{trk}}(\hat{\mathbf{X}}_t^{(l)},\,\mathbf{X}_t^{\text{obs}})
\Big)
\quad \text{s.t.}\quad
\hat{\mathbf{Z}}_{t+1}^{(l)} = F^{(l)}\!\big(\hat{\mathbf{Z}}_t^{(l)},\,\mathbf{u}_t^{(l)}\big),
$}
\label{eq:inv_problem}
\end{equation}
where $\mathcal{L}_{\text{geo}}$ and $\mathcal{L}_{\text{trk}}$ measure geometric and point-wise tracking discrepancies between the predicted positions $\hat{\mathbf{X}}_t^{(l)}$ and the observation $\mathbf{X}_t^{\text{obs}}$.

\begin{figure}[t]
\centering
\includegraphics[width=\textwidth]{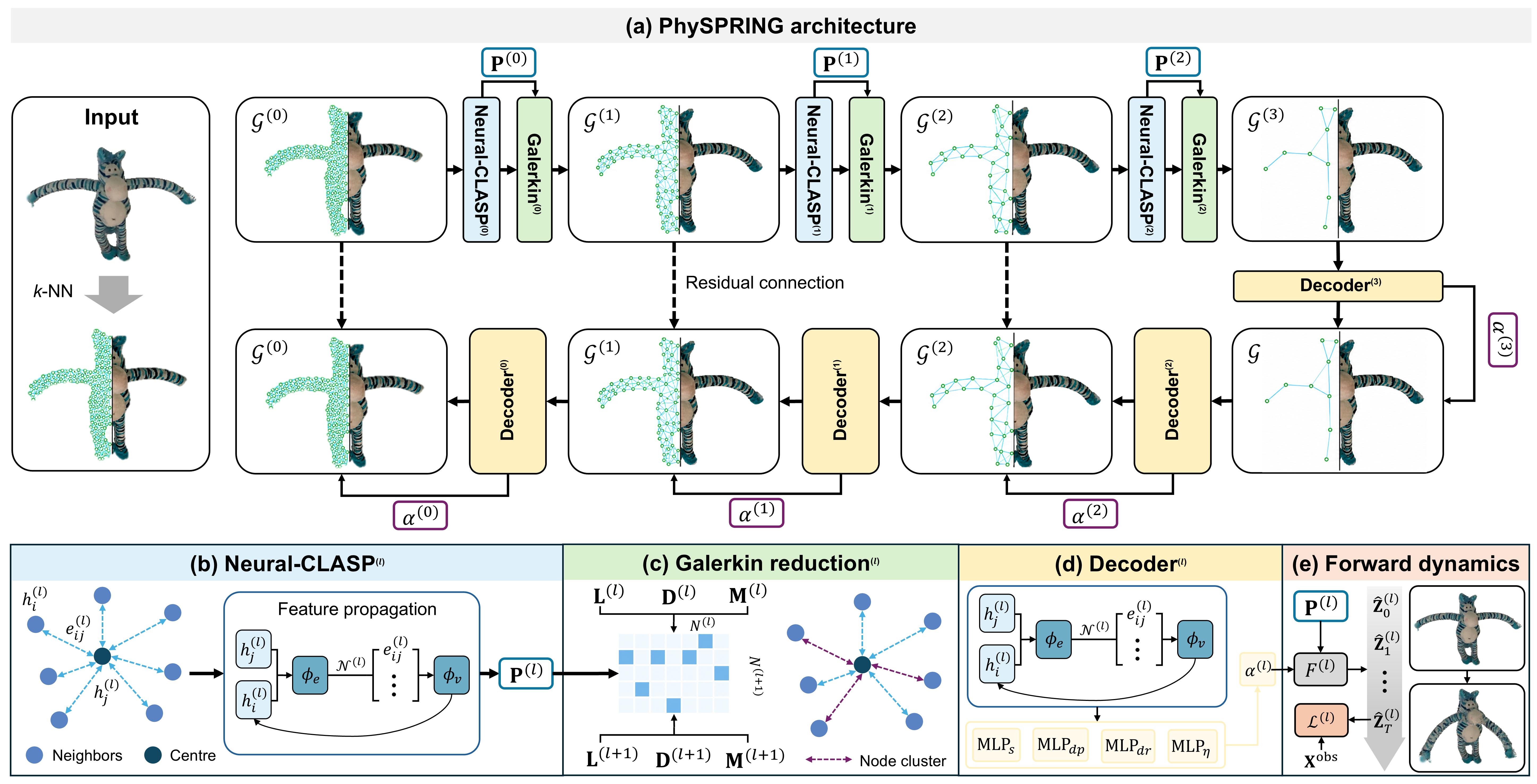}
\caption{Architecture of \method{}. \textbf{(a)}~U-Net-style hierarchical GNN to jointly optimize ROMs and the corresponding mechanical parameters $\alpha^{(l)}$ in different levels. \textbf{(b)} Neural-CLASP block produces $\mathbf{P}^{(l)}$. \textbf{(c)}~Galerkin reduction projects the level-$l$ system to their level-$(l{+}1)$ counterparts. \textbf{(d)}~Decoder shares the same message-passing structure as the encoder while outputting $\alpha^{(l)}$. \textbf{(e)}~Forward dynamics using numerical integrator $F^{(l)}$.}
\label{fig:method}
\end{figure}

\subsection{\method{} architecture}
\label{sec:spring_arch}

\method{} unifies the optimizations of $\{\mathbf{P}^{(l)}\}_{l=0}^{L-1}$ and $\{\alpha^{(l)}\}_{l=0}^{L}$ in a fully differentiable U-Net-style GNN method (Fig.~\ref{fig:method}a). The finest-level input graph $\mathcal{G}^{(0)}$ is initialized from a dense $k$-NN adjacency over the sampled particles from 3D reconstructed objects. The encoder then progressively coarsens $\mathcal{G}^{(0)}$ through $L$ (=3 in our paper) learnable neural clustering with adaptive structure preservation (Neural-CLASP) blocks, yielding a hierarchy $\mathcal{G}^{(0)}\succ\mathcal{G}^{(1)}\succ\dots\succ\mathcal{G}^{(L)}$ with monotonically decreasing node counts $N_0>N_1>\dots>N_L$. The
Neural-CLASP block at the end of each encoder level $l$ reads the encoder node features and outputs the assignment $\mathbf{P}^{(l)}$, which physically projects the level-$l$ spring--mass system onto the level-$(l{+}1)$ while preserving its second-order physical structure.

The decoder inherits the same per-level graph topology and message-passing structure from the encoder. Unlike the encoder, the decoder outputs $\alpha^{(l)}$ at every level $l$, while $\mathbf{M}^{(l)}$ is inherited deterministically through the Galerkin chain. Once $\mathbf{P}^{(l)}$ and $\alpha^{(l)}$ are produced, the forward dynamics are computed entirely through the
differentiable Euler integrator $F^{(l)}$ defined in Eq.~\eqref{eq:second_order}. This design guarantees that the level-$l$ ROM remains an explicit spring--mass system rather than a black-box GNN simulator, and is deployable in any compatible simulator for unseen interactions.

\subsubsection{Encoder and Neural-CLASP block}
As shown in Figure~\ref{fig:method}a and b, the Neural-CLASP block updates node and edge features through message-passing layers at each encoder level $l$. Unlike the classical second-order structure-preserving
MOR~\citep{model_reduction_2017}, \method{} directly learns $\mathbf{P}^{(l)}\in\{0,1\}$ from the dynamic and topological similarity of the GNN nodes rather than derive explicitly from a controllability Gramian~\citep{antoulas2005approximation}, therefore, relaxing the need for prior mechanical parameters.

Each node $i\in\mathcal{V}^{(0)}$ carries a time-invariant attribute tuple $(\mathbf{x}_i^{(0)}, m_i, \tau_i)$, where $\mathbf{x}_i^{(0)} \in \mathbf{X}_0^{(0)}$ and $\mathbf{x}_i^{(0)}\in\mathbb{R}^3$. $m_i \in \mathbf{M}^{(0)}$ and $\tau_i$ is a one-hot node-type indicator that distinguishes object material classes from boundary and controller classes. A high-frequency positional encoding $\Gamma(\cdot)$ adopted from NeRF~\citep{mildenhall2021nerf} is used to inject fine-scale geometry. The canonical node feature is therefore encoded as:
\begin{equation}
    \mathbf{h}_i^{(0)} = \mathrm{Enc}_v\!\big([\,\mathbf{x}_i^{(0)},\, m_i,\, \tau_i,\, \Gamma(\mathbf{x}_i^{(0)})\,]\big),
    \qquad i\in\mathcal{V}^{(0)},
\end{equation}
where $[\,\cdot\,]$ denotes concatenation along the feature axis and $\mathrm{Enc}_v$ is an MLP. Each candidate edge $(i,j)\in\mathcal{E}^{(0)}$ is initialized from a physics-aware geometric descriptor that captures the edge direction (force direction) and initial spring length,
\begin{equation}
    \mathbf{e}_{ij}^{(0)} = \mathrm{Enc}_e\!\big([\,\mathbf{x}_i^{(0)} - \mathbf{x}_j^{(0)},\, \|\mathbf{x}_i^{(0)} - \mathbf{x}_j^{(0)}\|\,]\big).
\end{equation}

\paragraph{Node and edge feature propagation.}
At every encoder level $l$, the edge feature $\mathbf{e}_{ij}^{(l)}$ is updated from the feature of the two endpoints (e.g., $\mathbf{h}_i^{(l)}$ and $\mathbf{h}_j^{(l)}$) and its previous state, and $\mathbf{h}_i^{(l)}$ is updated by aggregating the edge messages as follows:
\begin{equation}
    \mathbf{e}_{ij}^{(l)} \leftarrow \phi_e\!\big(\mathbf{h}_i^{(l)},\,\mathbf{h}_j^{(l)},\,\mathbf{e}_{ij}^{(l)},\,\tau_i,\,\tau_j\big),\quad
    \mathbf{h}_i^{(l)} \leftarrow \phi_v\!\Big(\mathbf{h}_i^{(l)},\,\textstyle\sum_{j\in\mathcal{N}^{(l)}(i)} \mathbf{e}_{ij}^{(l)}\Big),
    \label{eq:amp_update}
\end{equation}
where $\phi_e$ and $\phi_v$ are shared MLPs and $\mathcal{N}^{(l)}(i)$ defines the neighborhood of node $i$ in $\mathcal{G}^{(l)}$.

\paragraph{Discrete assignment via Gumbel--Softmax with STE.}
The Neural-CLASP block then measures the latent similarity between node pairs $i$ and $j$ via
$D_{ij}^{(l)} = \big\|\mathbf{h}_i^{(l)} - \mathbf{h}_j^{(l)}\big\|_{2}$, and converts the similarity scores $-D_{ij}^{(l)}$ into $\mathbf{P}^{(l)}$ through a Gumbel--Softmax
relaxation~\citep{jang2016categorical} integrated with a Straight-Through Estimator (STE)~\citep{bengio2013estimating},
\begin{equation}
    \tilde{P}_{ij}^{(l)} = \frac{\exp\!\big((-D_{ij}^{(l)} + g_{ij})/\lambda\big)}{\sum_{k\in\mathcal{V}^{(l)}}\exp\!\big((-D_{ik}^{(l)} + g_{ik})/\lambda\big)},\qquad
    P_{ij}^{(l)} =
    \begin{cases}
        1, & j = \arg\max_{k\in\mathcal{V}^{(l)}}\,\tilde{P}_{ik}^{(l)},\\
        0, & \text{otherwise,}
    \end{cases}
    \label{eq:gumbel_ste}
\end{equation}
where $g_{ij}\overset{\text{iid}}{\sim}\mathrm{Gumbel}(0,1)$ and $\lambda$
is the Gumbel temperature, and STE substitutes the hard $P_{ij}^{(l)}$ with the
soft $\tilde{P}_{ij}^{(l)}$ on the backward pass. The two relaxations enable the previous non-differentiable discrete topology selection of the spring--mass reduction in PhysTwin~\cite{jiang2025phystwin} into an end-to-end differentiable optimization problem.

\subsubsection{Galerkin reduction of system matrices}
Given $\mathbf{P}^{(l)}$, the $\mathbf{L}^{(l)}$ and $\mathbf{D}^{(l)}$ have the standard graph-Laplacian and damping-matrix structure:
\begin{equation}
    \resizebox{0.9\linewidth}{!}{$
    \displaystyle
    \mathbf{L}^{(l)}_{ij} =
    \begin{cases}
    -s_{ij}^{(l)}, & (i,j)\in\mathcal{E}^{(l)},\,i\neq j,\\[2pt]
    \sum_{k\in\mathcal{N}^{(l)}(i)} s_{ik}^{(l)}, & i=j,\\[2pt]
    0, & \text{otherwise,}
    \end{cases}\quad
    \mathbf{D}^{(l)}_{ij} =
    \begin{cases}
    -d_{\text{dp}}^{(l)}, & (i,j)\in\mathcal{E}^{(l)},\,i\neq j,\\[2pt]
    d_{\text{dr}}^{(l)} + |\mathcal{N}^{(l)}(i)|\,d_{\text{dp}}^{(l)}, & i=j,\\[2pt]
    0, & \text{otherwise.}
    \end{cases}
    $}
\end{equation}
where $s_{ij}^{(l)}$ is the per-edge spring stiffness, $d_{\text{dp}}^{(l)}$ a shared dashpot (relative-velocity) damping, and $d_{\text{dr}}^{(l)}$ a shared drag (absolute-velocity) damping.
As shown in Fig.~\ref{fig:method}c, the Galerkin projection is adopted by applying $\mathbf{P}^{(l)}$ to yield the coarsened spring--mass system at level $l+1$:
\begin{equation}
    \hat{\mathbf{M}}^{(l+1)} = (\mathbf{P}^{(l)})^{\top}\mathbf{M}^{(l)}\mathbf{P}^{(l)},
    \quad
    \hat{\mathbf{L}}^{(l+1)} = (\mathbf{P}^{(l)})^{\top}\mathbf{L}^{(l)}\mathbf{P}^{(l)},
    \quad
    \hat{\mathbf{D}}^{(l+1)} = (\mathbf{P}^{(l)})^{\top}\mathbf{D}^{(l)}\mathbf{P}^{(l)},
\label{eq:galerkin_proj}
\end{equation}
The mass projection is deterministic, so $\mathbf{M}^{(l+1)}$ is fully determined by $\mathbf{P}^{(l)}$ and the fixed $\mathbf{M}^{(0)}$, while $\hat{\mathbf{L}}^{(l+1)}$ and $\hat{\mathbf{D}}^{(l+1)}$ serve as the initial estimations on mechanical parameters, where the decoder later corrects through a learnable bias. The $\hat{\mathcal{E}}^{(l+1)}$ is inherited from the off-diagonal sparsity of $\hat{\mathbf{L}}^{(l+1)}$.

\subsubsection{Decoder and parameter heads}
The decoder reuses the per-level ROMs directly from the encoder while applying in reverse (e.g., $l = L \to 0$) with residual skip connections (Fig.~\ref{fig:method}d). Node and edge feature update in the decoder follows the same rules as in the encoder. The decoder differs only at the output, where three MLP layers (e.g., $\mathrm{MLP}_s$, $\mathrm{MLP}_{\text{dp}}$, $\mathrm{MLP}_{\text{dr}}$) read $\mathbf{e}_{ij}^{(l)}$ and predict residual corrections to the initial mechanical parameter estimations from the Galerkin projection. Summing each correction with its base yields the level-$l$ mechanical parameters:
\begin{equation}
s_{ij}^{(l)} = \hat{s}_{ij}^{(l)} + \beta_s\,\mathrm{MLP}_s(\mathbf{e}_{ij}^{(l)}),\;\;
d_{\text{dp}}^{(l)} = \hat{d}_{\text{dp}}^{(l)} + \beta_d\,\mathrm{MLP}_{\text{dp}}^{(l)},\;\;
d_{\text{dr}}^{(l)} = \hat{d}_{\text{dr}}^{(l)} + \beta_d\,\mathrm{MLP}_{\text{dr}}^{(l)},
\label{eq:bias}
\end{equation}
A lightweight contact head $\mathrm{MLP}_{\eta}$ directly predicts the contact coefficients $\boldsymbol{\eta}^{(l)}$. 

\paragraph{Forward dynamics and training loss.}
As shown in Fig.~\ref{fig:method}e, the predicted $\alpha^{(l)}$ and $\mathcal{G}^{(l)}$ are fed to $F^{(l)}$ (e.g., Eq.~\eqref{eq:second_order}) to roll out the dynamical state recursively from the projected initial condition $\hat{\mathbf{Z}}_0^{(l)}$ along the full observation sequence, as shown in the following equation:
\begin{equation}
    \hat{\mathbf{Z}}_{t+1}^{(l)} = F^{(l)}\!\big(\hat{\mathbf{Z}}_t^{(l)},\,\mathbf{u}_t^{(l)};\,\mathbf{M}^{(l)},\,\alpha^{(l)},\,\mathcal{G}^{(l)}\big), \qquad t = 0, 1, \ldots, T-1.
    \label{eq:rollout}
\end{equation}
This is the key architectural difference between \method{} and EvoMesh~\citep{evomesh_icml2024}, where Evomesh relies on a black-box GNN to learn to predict the future dynamics rather than directly modeling the entire physical system. The training loss averages the per-frame $\mathcal{L}_{\text{geo}}$ and $\mathcal{L}_{\text{trk}}$ along the whole trajectory and sums across all levels,
\begin{equation}
    \mathcal{L} \;=\; \sum_{l=0}^{L} \frac{1}{T}\sum_{t=1}^{T}
    \Big[
    \mathcal{L}_{\text{geo}}\!\big(\hat{\mathbf{X}}_t^{(l)},\,\mathbf{X}_t^{\text{obs}}\big)
    +
    \mathcal{L}_{\text{trk}}\!\big(\hat{\mathbf{X}}_t^{(l)},\,\mathbf{X}_t^{\text{obs}}\big)
    \Big],
    \label{eq:training_loss}
\end{equation}
and the gradient $\nabla_{\Theta}\mathcal{L}$ propagates back through the full differentiable rollout to update the network weights $\Theta$. Crucially, this single forward--loss--backward pass jointly optimizes both $\{\mathbf{P}^{(l)}\}$ and $\{\alpha^{(l)}\}$ in one stage (in contrast to the two-stage pipeline of PhysTwin~\citep{jiang2025phystwin}). The progressive training strategy that schedules this joint optimization through Gumbel-temperature annealing is described in Appendix~\ref{app:training_details}.

\section{Experiments}
\label{sec:experiments}

We evaluate \method{} on the PhysTwin~\citep{jiang2025phystwin} and Real2Sim~\citep{2026arXiv260301151L} benchmarks to demonstrate three main contributions stated in the introduction.

\subsection{Experimental setup}
\label{sec:exp_setup}

\textbf{Baselines.} 
PhysTwin~\cite{jiang2025phystwin} reconstructs a spring-mass digital twin from RGB-D video through a two-stage pipeline. We use PhysTwin benchmark to validate the 4D reconstruction and prediction quality of \method{} from FOM and ROMs. We follow the original train/test split in Phystwin. Real2Sim~\citep{2025arXiv251104665Z} deploys deformable digital twins inside a robot policy-evaluation simulator, where the real-to-sim gaps are validated on rope object reconstructed via PhysTwin and evaluated on downstream manipulation tasks. We re-train \method{} on the rope to obtain ROMs, and substitute them into the same Real2Sim simulation environment without retraining the policy sampling networks to assess the impact of order reduction on downstream policy execution.

\textbf{Metrics.}
We adopt the same metrics used in PhysTwin. For 3D evaluation, Chamfer Distance (CD) and Tracking Error computed from rollouts of the reconstructed spring--mass system. For 2D evaluation, PSNR, SSIM, LPIPS, and IoU computed from 3DGS renderings of the reconstructed twin. For Real2Sim downstream evaluation, we adopt the same success-ratio measurement as Real2Sim over the same imitation learning policies, i.e., ACT~\citep{zhao2023learning} and $\pi_0$~\citep{black2024pi_0}, alongside inference speed to quantify the speed-up gained from order reduction.

\subsection{Reconstruction\&re-simulation and prediction across ROMs}
\label{sec:exp_recon}
\begin{figure}[th!]
\centering
\includegraphics[width=\textwidth]{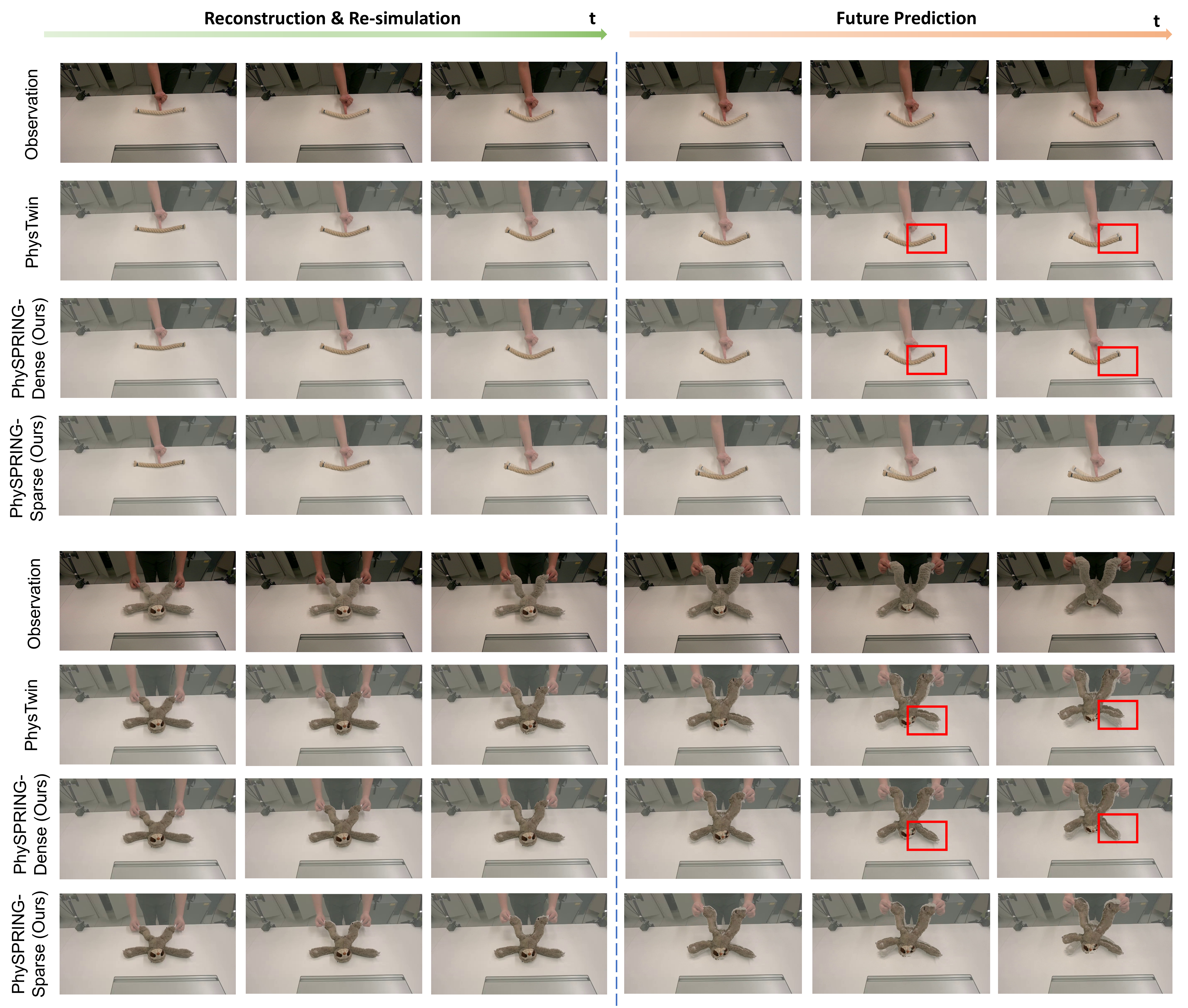}
\caption{The qualitative results of \method{} and the comparison method on the reconstruction \& re-simulation and future prediction of the object deformation, visualized only at the FOM representation.}
\label{fig:recon_pred}
\end{figure}

We first evaluate reconstruction\&re-simulation and future prediction on the PhysTwin benchmark. As shown in Table~\ref{tab:t1_recon_pred}, \method{} achieves an improvement on PhysTwin at the dense Level-0 representation. For reconstruction\&re-simulation, \method{}  reduces CD by 17.0\% and tracking error by 11.1\%. For future prediction, it reduces CD and tracking error by 17.2\% and 10.5\%, respectively. \method{}  also improves rendering quality, where PSNR increases by 0.70 dB for reconstruction and 0.62 dB for prediction, and LPIPS decreases by 17.6\% and 10.9\%, respectively. Qualitative comparisons further show that \method{} better captures object deformation and temporal alignment, producing reconstructions that more closely match the observed dynamics (see Fig.~\ref{fig:recon_pred} red insets).

\begin{table*}[t]
\centering
\caption{Quantitative results on reconstruction \& re-simulation and future prediction. Ratio reports per-level compression, and FPS reports effective across all frames.}
\label{tab:t1_recon_pred}
\resizebox{\textwidth}{!}{%
\begin{tabular}{l|cc|cccccc|cccccc}
\toprule
& & & \multicolumn{6}{c|}{Reconstruction \& Re-simulation} & \multicolumn{6}{c}{Future Prediction} \\
\cline{4-15}
Method & Ratio & FPS $\uparrow$
& CD $\downarrow$ & Track $\downarrow$ & IoU\,\% $\uparrow$ & PSNR $\uparrow$ & SSIM $\uparrow$ & LPIPS $\downarrow$
& CD $\downarrow$ & Track $\downarrow$ & IoU\,\% $\uparrow$ & PSNR $\uparrow$ & SSIM $\uparrow$ & LPIPS $\downarrow$ \\
\midrule
Spring-Gaus~\citep{zhong2024reconstruction} & 1$\times$ & / & 0.041 & 0.050 & 57.6 & 23.445 & 0.928 & 0.102 
& 0.062 & 0.094 & 46.4 & 22.488 & 0.924 & 0.113   \\
GS-Dynamics~\citep{zhang2024dynamics} & 1$\times$ & / & 0.014 & 0.022 & 72.1 & 26.260 & 0.940 & 0.052 
& 0.041 & 0.070 & 49.8 & 22.540 & 0.924 & 0.097 \\
PhysTwin~\citep{jiang2025phystwin} & 1$\times$ & 2.3 & 0.0053 & 0.0090 & 84.40 & 28.214 & 0.945 & 0.034 & 0.0116 & 0.0220 & 72.50 & 25.617 & 0.941 & 0.055 \\
\textbf{Ours (Level 0)}  & 1$\times$ & 2.3 & \textbf{0.0044} & \textbf{0.0080} & \textbf{85.34} & \textbf{28.91} & \textbf{0.950} & \textbf{0.028} & \textbf{0.0096} & \textbf{0.0197} & \textbf{72.61} & \textbf{26.238} & \textbf{0.945} & \textbf{0.049} \\
\midrule
\textbf{Ours (Level 1)}  & 60\% & 3.8 & 0.0069 & 0.0111 & 81.45 & 28.082 & 0.947 & 0.034 & 0.0127 & 0.0244 & 65.90 & 25.000 & 0.939 & 0.061 \\
\textbf{Ours (Level 2)}  & 40\% & 4.6 & 0.0091 & 0.0127 & 80.09 & 27.718 & 0.946 & 0.037 & 0.0158 & 0.0281 & 64.15 & 24.764 & 0.937 & 0.066 \\
\textbf{Ours (Level 3)}  & 30\% & 5.3 & 0.0120 & 0.0161 & 77.74 & 27.219 & 0.944 & 0.041 & 0.0201 & 0.0337 & 60.44 & 24.060 & 0.935 & 0.071 \\
\bottomrule
\end{tabular}}
\end{table*}

\begin{wraptable}{l}{0.6\textwidth}
\vspace{-8pt}
\centering
\caption{Fidelity--efficiency trade-off across ROM levels.}
\label{tab:rom_tradeoff}
\scriptsize
\setlength{\tabcolsep}{2.5pt}
\renewcommand{\arraystretch}{1.05}
\resizebox{\linewidth}{!}{%
\begin{tabular}{lccccc}
\toprule
ROM & \makecell{Ratio} & \makecell{System\\Reduction} 
& \makecell{Re\&sim\\CD/Track} & \makecell{Pred.\\CD/Track} 
& \makecell{FPS /\\Speed-up} \\
\midrule
Level 1 & 60\% & 1.67$\times$ & 1.57/1.39$\times$ & 1.32/1.24$\times$ & 3.8 / 1.65$\times$ \\
Level 2 & 40\% & 2.50$\times$ & 2.07/1.59$\times$ & 1.65/1.43$\times$ & 4.6 / 2.00$\times$ \\
Level 3 & 30\% & 3.33$\times$ & 2.73/2.01$\times$ & 2.09/1.71$\times$ & 5.3 / 2.30$\times$ \\
\bottomrule
\end{tabular}%
}
\vspace{-10pt}
\end{wraptable}
The physical accuracy--efficiency trade-off of progressive reduction is summarized in Table~\ref{tab:rom_tradeoff}. Across Levels~1--3, the 3D error increases remain smaller than the corresponding reductions in system size, indicating that \method{} preserves the dominant deformation dynamics even with fewer simulated points. At the same time, wall-clock throughput improves monotonically from Level~0 to Level~3, reaching up to a 2.30$\times$ speed-up. Since the 2D rendering quality degrades only mildly, these results show that \method{} achieves meaningful efficiency gains while maintaining stable physical and visual fidelity.



\subsection{Downstream Real2Sim policy evaluation}
\label{sec:exp_real2sim}
\begin{figure}[t]
\centering
\begin{minipage}[t]{0.32\textwidth}
\centering
\includegraphics[width=\linewidth]{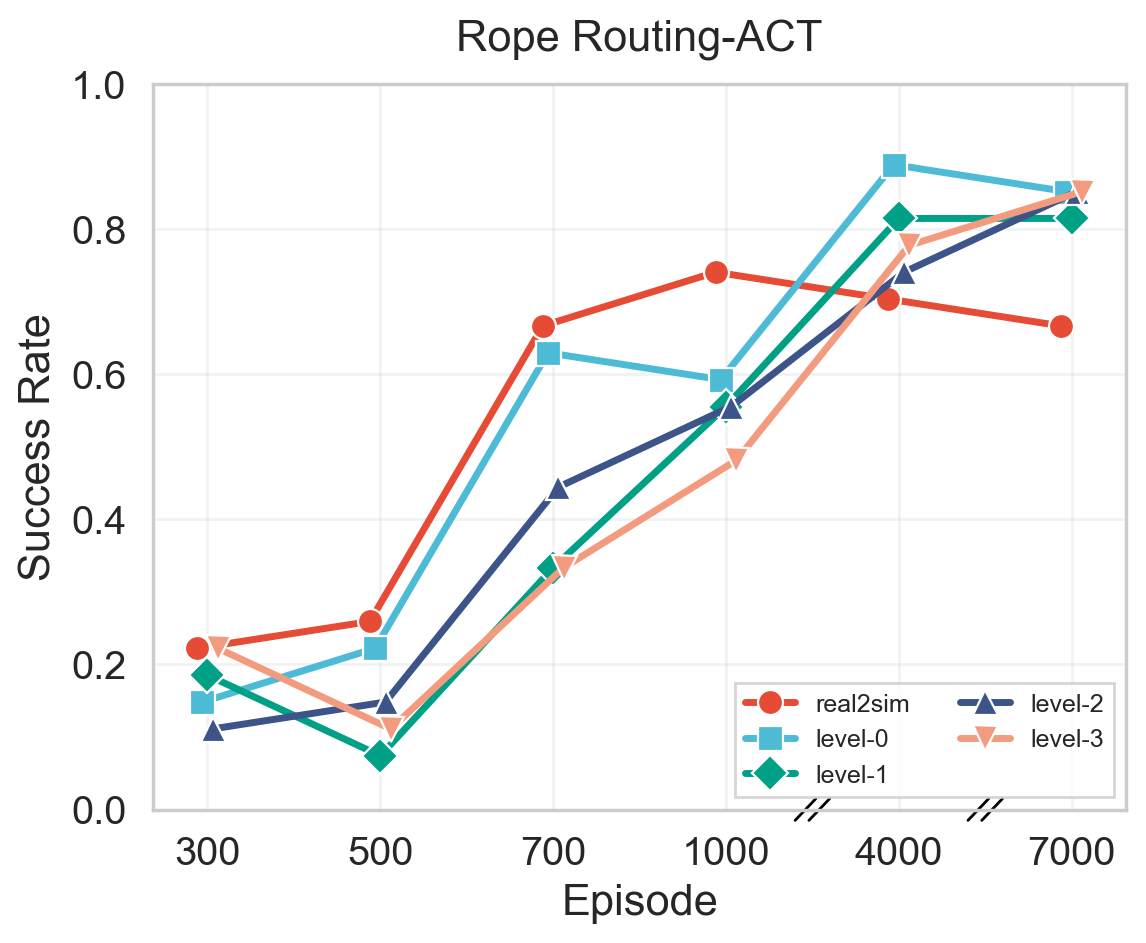}
\par\vspace{1mm}
{\small \textbf{(a)} ACT success rate\par}
\end{minipage}\hfill
\begin{minipage}[t]{0.32\textwidth}
\centering
\includegraphics[width=\linewidth]{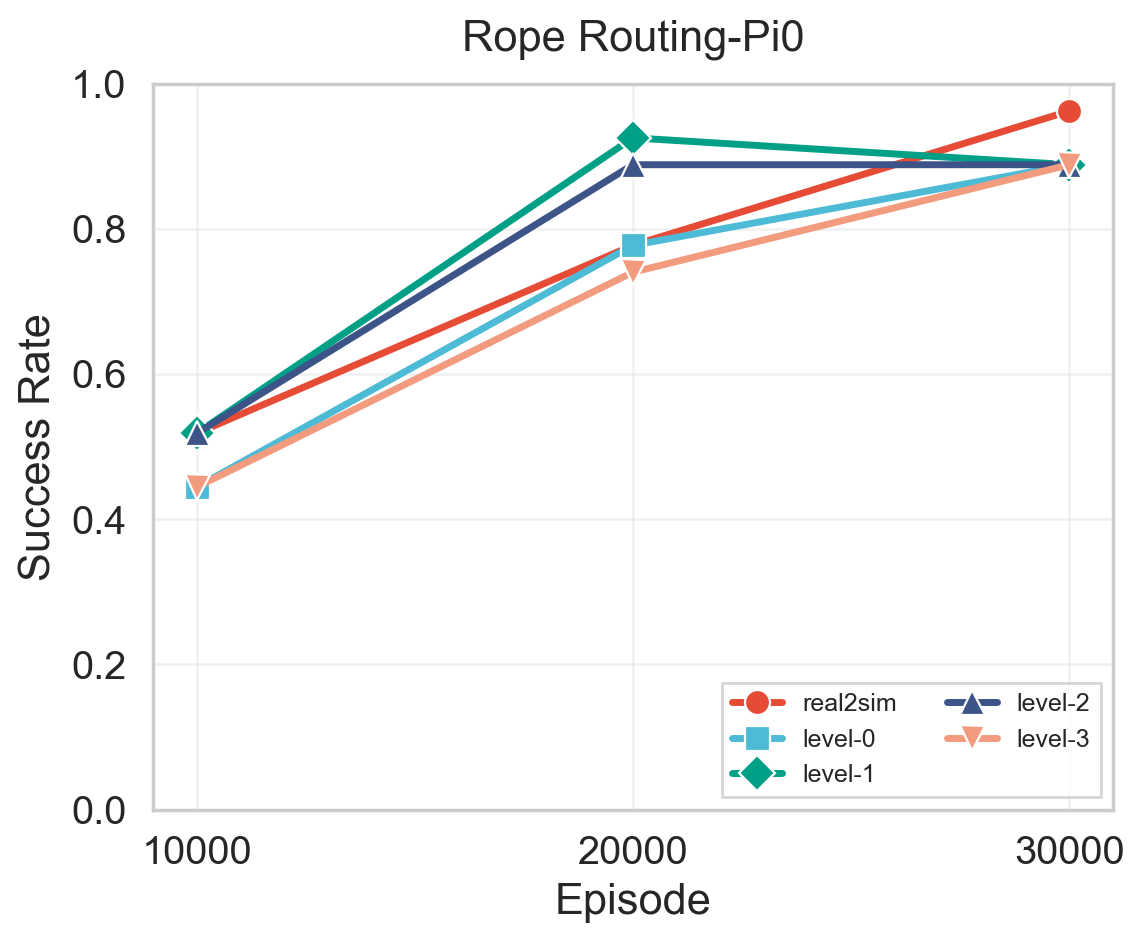}
\par\vspace{1mm}
{\small \textbf{(b)} $\pi_0$ success rate\par}
\end{minipage}\hfill
\begin{minipage}[t]{0.32\textwidth}
\centering
\includegraphics[width=\linewidth]{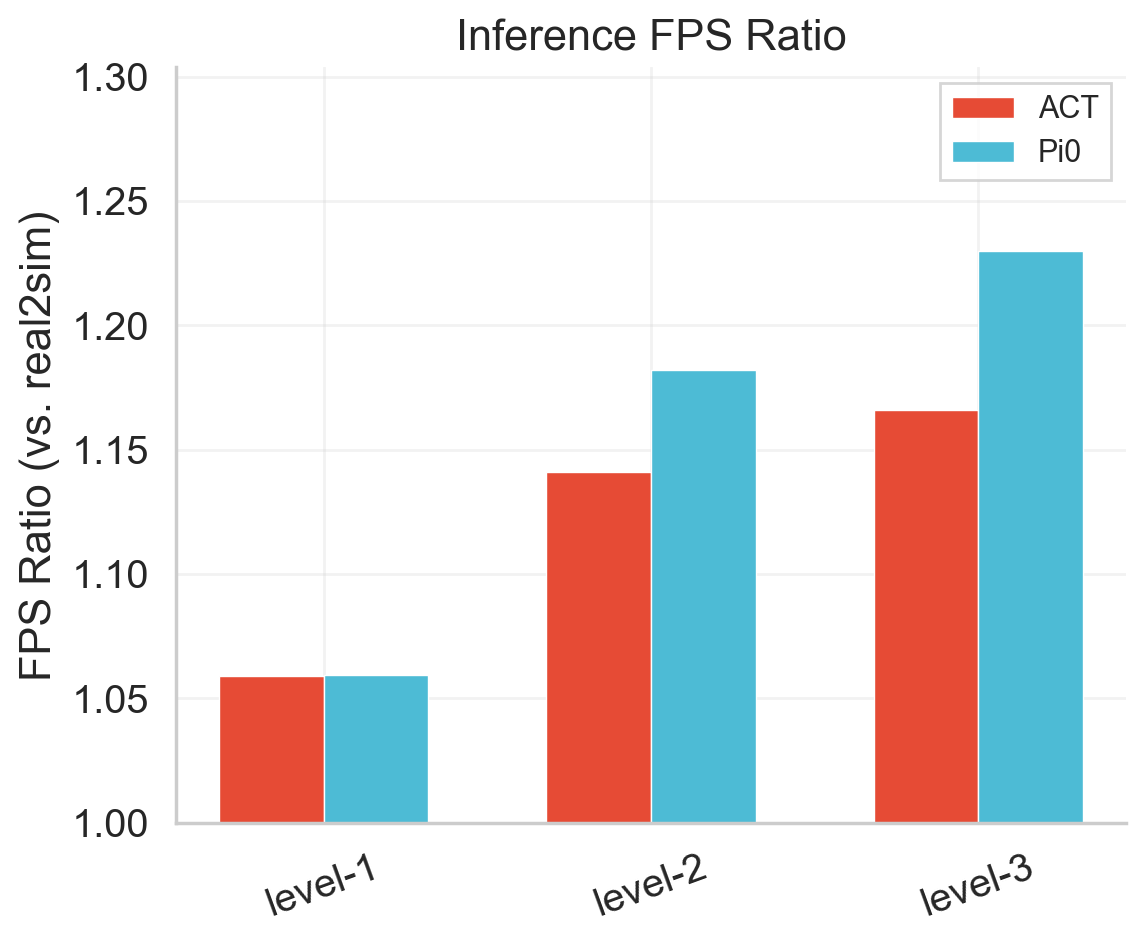}
\par\vspace{1mm}
{\small \textbf{(c)} Action-sampling effectiveness\par}
\end{minipage}
\caption{Zero-shot Real2Sim policy evaluation across ROM levels.}
\label{fig:real2sim_zero_shot}
\end{figure}

We deploy each ROM trained by \method{} inside the Real2Sim~\citep{2025arXiv251104665Z} policy-evaluation simulator on the rope-routing task, and quantify the impact of order reduction on downstream policy execution using ACT~\citep{zhao2023learning} and $\pi_0$~\citep{black2024pi_0} under the original Real2Sim setup. We evaluate the ROMs in a zero-shot setting by directly substituting each \method{} ROM into the same simulator and rerunning the policy-evaluation loop without further training or adaptation.
As shown in Fig.~\ref{fig:real2sim_zero_shot}a and b, both policies maintain comparable success rates across all ROM levels, indicating that the reduced models preserve the task-relevant contact and deformation dynamics required for policy evaluation. Since the policy input is image-based, the visual fidelity of the ROMs also remains sufficient for the policies to recognize the task state and output effective actions. At the same time, progressive reduction improves action-sampling effectiveness, achieving maximum effective gains of 17\% for ACT and 23\% for $\pi_0$ while maintaining policy success, as shown in Fig.~\ref{fig:real2sim_zero_shot}c.
The full success-count table across policy-training iterations, per-step inference timing, and qualitative motion-sequence comparisons across ROM levels are reported in Appendices~\ref{app:real2sim_success}, \ref{app:real2sim_timing}, and \ref{app:real2sim_motion}, respectively.

\subsection{Ablations}
\label{sec:exp_ablation}

\paragraph{Effectiveness of Neural-CLASP} To isolate the contribution of Neural-CLASP, we compare \method{} with two alternative reduction strategies, i.e., random sampling and classical Gramian-based MOR~\citep{model_reduction_2017}. All variants use the same \method{} backbone and training settings, where only the node-reduction mechanism is changed. As shown in Table~\ref{tab:t3_sampling}, the learned Neural-CLASP sampler consistently outperforms both baselines at matched node budgets. Compared with random sampling, Neural-CLASP better preserves dynamically important and geometrically representative regions, leading to lower 3D errors and improved rendering quality for both reconstruction\&re-simulation and future prediction. It also improves over the Gramian baseline, whose fixed analytic reduction is less adaptive to observation-supervised geometry. These results show that Neural-CLASP learns a reduced topology that better retains the mechanically active structure needed by the downstream simulator.

\begin{table*}[t]
\centering
\caption{Sampling-strategy ablation under same sampling rate.}
\label{tab:t3_sampling}
\resizebox{\textwidth}{!}{%
\begin{tabular}{l|c|cccccc|cccccc}
\toprule
& & \multicolumn{6}{c|}{Reconstruction \& Re-simulation} & \multicolumn{6}{c}{Future Prediction} \\
\cline{3-14}
Method & Ratio
& CD $\downarrow$ & Track $\downarrow$ & IoU\,\% $\uparrow$ & PSNR $\uparrow$ & SSIM $\uparrow$ & LPIPS $\downarrow$
& CD $\downarrow$ & Track $\downarrow$ & IoU\,\% $\uparrow$ & PSNR $\uparrow$ & SSIM $\uparrow$ & LPIPS $\downarrow$ \\
\midrule
Random                       & 60\% & 0.0112 & 0.0177 & 74.71 & 26.946 & 0.939 & 0.046 & 0.0246 & 0.0446 & 54.85 & 23.405 & 0.922 & 0.088 \\
Random                       & 40\% & 0.0183 & 0.0275 & 69.78 & 26.325 & 0.941 & 0.051 & 0.0493 & 0.0819 & 41.55 & 21.977 & 0.919 & 0.108 \\
Random                       & 30\% & 0.0133 & 0.0182 & 76.34 & 26.988 & 0.943 & 0.043 & 0.0203 & 0.0377 & 58.15 & 23.769 & 0.930 & 0.078 \\
\midrule
Gramian (classical)          & 60\% & 0.0078 & 0.0124 & 80.26 & 27.788 & 0.946 & 0.037 & 0.0163 & 0.0301 & 62.88 & 24.525 & 0.935 & 0.069 \\
Gramian (classical)          & 40\% & 0.0093 & 0.0139 & 79.61 & 27.710 & 0.945 & 0.038 & 0.0165 & 0.0316 & 62.41 & 24.608 & 0.936 & 0.068 \\
Gramian (classical)          & 30\% & 0.0138 & 0.0186 & 77.12 & 27.138 & 0.944 & 0.041 & 0.0230 & 0.0385 & 59.36 & 24.006 & 0.933 & 0.074 \\
\midrule
\textbf{Learnable (Ours)}    & 60\% & \textbf{0.0069} & \textbf{0.0111} & \textbf{81.45} & \textbf{28.082} & \textbf{0.947} & \textbf{0.034} & \textbf{0.0127} & \textbf{0.0244} & \textbf{65.90} & \textbf{25.000} & \textbf{0.939} & \textbf{0.061} \\
\textbf{Learnable (Ours)}    & 40\% & \textbf{0.0091} & \textbf{0.0127} & \textbf{80.09} & \textbf{27.718} & \textbf{0.946} & \textbf{0.037} & \textbf{0.0158} & \textbf{0.0281} & \textbf{64.15} & \textbf{24.764} & \textbf{0.937} & \textbf{0.066} \\
\textbf{Learnable (Ours)}    & 30\% & \textbf{0.0120} & \textbf{0.0161} & \textbf{77.74} & \textbf{27.219} & 0.944 & 0.041 & \textbf{0.0201} & \textbf{0.0337} & \textbf{60.44} & \textbf{24.060} & \textbf{0.935} & \textbf{0.071} \\
\bottomrule
\end{tabular}}
\end{table*}

\paragraph{End-to-end ROM training.} We further compare \method{} with a PhysTwin-style two-stage training pipeline across different reduction ratios, where we use random node sampling on PhysTwin. As shown in Table~\ref{tab:t4_phystwin_train}, \method{}  consistently outperforms the PhysTwin pipeline at all sampling ratios for both reconstruction\&re-simulation and future prediction. The improvement is substantial across both 2D and 3D metrics, and remains stable as the reduction ratio becomes more substantial. These results indicate that the end-to-end \method{} architecture, which jointly learns reduction and dynamics instead of separating them into Stage 0 and Stage 1, produces more accurate and robust ROMs both in dense and sparse levels.

\begin{table*}[t]
\centering
\caption{Comparison against the PhysTwin training pipeline across ROMs.}
\label{tab:t4_phystwin_train}
\resizebox{\textwidth}{!}{%
\begin{tabular}{l|c|cccccc|cccccc}
\toprule
& & \multicolumn{6}{c|}{Reconstruction \& Re-simulation} & \multicolumn{6}{c}{Future Prediction} \\
\cline{3-14}
Method & Ratio
& CD $\downarrow$ & Track $\downarrow$ & IoU\,\% $\uparrow$ & PSNR $\uparrow$ & SSIM $\uparrow$ & LPIPS $\downarrow$
& CD $\downarrow$ & Track $\downarrow$ & IoU\,\% $\uparrow$ & PSNR $\uparrow$ & SSIM $\uparrow$ & LPIPS $\downarrow$ \\
\midrule
PhysTwin (Level 1)            & 60\% & 0.0542 & 0.0600 & 58.86 & 24.145 & 0.919 & 0.085 & 0.0792 & 0.0895 & 42.98 & 21.701 & 0.907 & 0.112 \\
PhysTwin (Level 2)            & 40\% & 0.0647 & 0.0879 & 53.32 & 23.904 & 0.920 & 0.091 & 0.0974 & 0.1281 & 37.10 & 21.523 & 0.906 & 0.119 \\
PhysTwin (Level 3)            & 30\% & 0.0810 & 0.1074 & 53.51 & 23.689 & 0.917 & 0.091 & 0.1366 & 0.1690 & 33.66 & 21.411 & 0.906 & 0.120 \\
\midrule
\textbf{Ours (Level 1)}      & 60\% & \textbf{0.0069} & \textbf{0.0111} & \textbf{81.45} & \textbf{28.082} & \textbf{0.947} & \textbf{0.034} & \textbf{0.0127} & \textbf{0.0244} & \textbf{65.90} & \textbf{25.000} & \textbf{0.939} & \textbf{0.061} \\
\textbf{Ours (Level 2)}      & 40\% & \textbf{0.0091} & \textbf{0.0127} & \textbf{80.09} & \textbf{27.718} & \textbf{0.946} & \textbf{0.037} & \textbf{0.0158} & \textbf{0.0281} & \textbf{64.15} & \textbf{24.764} & \textbf{0.937} & \textbf{0.066} \\
\textbf{Ours (Level 3)}      & 30\% & \textbf{0.0120} & \textbf{0.0161} & \textbf{77.74} & \textbf{27.219} & \textbf{0.944} & \textbf{0.041} & \textbf{0.0201} & \textbf{0.0337} & \textbf{60.44} & \textbf{24.060} & \textbf{0.935} & \textbf{0.071} \\
\bottomrule
\end{tabular}}
\end{table*}

\section{Conclusion and limitations}
\label{sec:conclusion_limitations}

We introduced \method{}, an end-to-end GNN-based framework for structure-preserving reduction of spring--mass digital twins reconstructed from observations. Unlike black-box neural simulators, \method{} preserves every ROM as an explicit second-order physical system by enabling differentiable topology reduction and jointly learning mechanical parameters across reduction levels. Experiments on PhysTwin show that our dense model improves reconstruction\&simulation and future-prediction accuracy over PhysTwin, while the ROMs provide a controlled fidelity--efficiency trade-off with up to a 2.30$\times$ simulation speed-up. In downstream Real2Sim policy evaluation, the same ROMs can be substituted into ACT and $\pi_0$ evaluations in zero-shot manner, preserving comparable manipulation success rates in rope-routing tasks while improving action-sampling effectiveness by up to 1.23$\times$. These results suggest that learned physical reduction can make reconstructed digital twins more practical for repeated robot-interaction rollouts without discarding their simulator-compatible structure.

\paragraph{Limitations.}
Two limitations remain. First, the improvement in policy-sampling effectiveness mainly comes from faster forward-dynamics prediction with ROMs, while the policy-sampling network itself provides limited additional acceleration. Future work should explore simpler physical-representation-driven models for more efficient policy sampling using ROMs. Second, training is slower than the original PhysTwin pipeline because progressive topology training waits for higher-resolution structures to stabilize before continuing to lower-resolution levels, introducing additional computational cost. More robust and efficient training schedules are needed.

{
\small
\bibliographystyle{unsrt}
\bibliography{references}
}


\newpage
\appendix



\section{Implementation details}
\label{app:training_details}

\paragraph{Training details.}
The active configuration uses $L = 3$ (comprising four levels, including $l=0$); latent dimension $128$; positional-encoding dimension $30$. Training runs $E = 80$ epochs. Optimization uses Adam with learning rate  $lr = 1e-3$ and decay $\gamma_{\text{lr}} = 0.9$. The differentiable Euler integrator runs $667$ Warp substeps per simulation step.

\method{} is implemented using the PyTorch framework, and the spring--mass physical simulation is built on the Warp GPU-accelerated simulator~\citep{macklin2022warp}. To enable end-to-end training from the Warp simulator to the PyTorch network, we bridge the gradient information computed inside Warp back to the PyTorch computational graph, allowing losses defined on Warp simulation rollouts to directly optimize the GNN parameters. We perform five levels of downsampling in the encoder; however, only the first four levels ($l = 0,\dots,3$) are passed to the Warp simulator the optimization. All experiments are conducted on a single NVIDIA RTX 4090 GPU.
\begin{algorithm}[H]
\caption{\method{} training with progressive topology commitment.}
\label{alg:spring_train}
\KwIn{Observations $\{\mathbf{X}_t^{\text{obs}}\}_{t=0}^{T}$, manipulator inputs $\{\mathbf{u}_t\}_{t=0}^{T}$, FOM graph $\mathcal{G}^{(0)}$, mass $\mathbf{M}^{(0)}$, levels $L$, epochs $E$, commit interval $E_{\text{commit}}$, model updates $K_{\text{model}}$, collision updates $K_{\text{col}}$}
\KwOut{Trained parameters $\Theta$, hierarchy $\{\mathbf{P}^{(l)}, \alpha^{(l)}\}_{l=0}^{L}$}
$\Theta \leftarrow \mathrm{init}$\;
$\mathbf{P}_{\text{stored}}^{(l)} \leftarrow \emptyset\;\forall\,l$\;
\For{$\mathrm{epoch} = 1$ \KwTo $E$}{
    $l^* \leftarrow \lfloor \mathrm{epoch} \,/\, E_{\text{commit}} \rfloor - 1$ \tcp*{level supervised this epoch}
    \For{$\mathrm{iter} = 1$ \KwTo $K_{\text{model}}$}{
        \tcp{Encoder pass (downward): message-passing, topology learning, coarsening}
        \For{$l = 0$ \KwTo $L-1$}{
            update $\{\mathbf{h}^{(l)}, \mathbf{e}^{(l)}\}$ via message-passing on $\mathcal{G}^{(l)}$\;
            \eIf{$l \le l^*$}{
                $\mathbf{P}^{(l)} \leftarrow \mathbf{P}_{\text{stored}}^{(l)}$ \tcp*{topology committed}
            }{
                $\mathbf{P}^{(l)} \leftarrow \mathrm{Neural\text{-}CLASP}\!\big(\mathbf{h}^{(l)}\big)$ \tcp*{Gumbel--Softmax + STE}
                $\mathbf{P}_{\text{stored}}^{(l)} \leftarrow \mathbf{P}^{(l)}$\;
            }
            $\big(\mathbf{M}^{(l+1)}, \mathcal{G}^{(l+1)}\big) \leftarrow \mathrm{GalerkinProject}\!\big(\mathbf{M}^{(l)}, \mathcal{G}^{(l)};\,\mathbf{P}^{(l)}\big)$ \tcp*{Eq.~\eqref{eq:galerkin_proj}}
        }
        \tcp{Decoder pass (upward): decode from coarse to fine, supervise at $l^*$}
        \For{$l = L-1$ \textbf{down to } $l^*$}{
            decode features at level $l$ via message-passing\;
            $\alpha^{(l)} = \big(\mathbf{L}^{(l)}, \mathbf{D}^{(l)}, \boldsymbol{\eta}^{(l)}\big)$ \tcp*{decode stiffness, damping via Eq.~\eqref{eq:bias}}
            \If{$l = l^*$}{
                $\hat{\mathbf{Z}}_{t+1}^{(l)} = F^{(l)}\!\big(\hat{\mathbf{Z}}_t^{(l)},\,\mathbf{u}_t^{(l)};\,\mathbf{M}^{(l)},\,\alpha^{(l)},\,\mathcal{G}^{(l)}\big)$,\, $t = 0, \ldots, T-1$\;
                $\mathcal{L} \leftarrow \frac{1}{T}\sum_{t=1}^{T}\!\big[\mathcal{L}_{\text{geo}} + \mathcal{L}_{\text{trk}}\big]\!\big(\hat{\mathbf{X}}_t^{(l)},\,\mathbf{X}_t^{\text{obs}}\big)$\;
                $\Theta \leftarrow \mathrm{Adam}\!\big(\Theta,\,\nabla_{\Theta}\mathcal{L}\big)$\;
            }
        }
    }
    \tcp{Collision parameter refinement}
    \For{$\mathrm{iter} = 1$ \KwTo $K_{\text{col}}$}{
        refine collision parameters at level $l^*$\;
    }
}
\Return{$\Theta$,\, $\{\mathbf{P}^{(l)}, \alpha^{(l)}\}_{l=0}^{L}$}
\end{algorithm}

\paragraph{Progressive topology training.}
The topology of each level $\mathcal{G}^{(l)}$ is determined by the model prediction $\mathbf{P}^{(l)}$, and therefore changes throughout training. As a result, even small perturbations in $\mathbf{P}^{(l)}$ can lead to large variations in $\alpha^{(l)}$, making the overall optimization unstable. To mitigate this issue, we progressively freeze the topology of each level during training. Once a level is fixed, corresponding $\mathbf{P}^{(l)}$ stops updating, while lower levels remain learnable and continue to optimize their downsampling operators. During backpropagation, we only supervise predictions from the upper levels whose topologies have already been committed, thereby decoupling the learning of mechanical parameters from the ongoing topology search in deeper layers. This strategy stabilizes training while preserving dynamic, learnable downsampling topology within a single optimization of $\mathcal{L}(\Theta)$. Algorithm~\ref{alg:spring_train} summarizes the resulting training loop.

\section{Additional results}

\subsection{Qualitative visualization of reconstruction\&re-simulation and prediction across ROMs}

\begin{figure}[t]
\centering
\includegraphics[width=\textwidth]{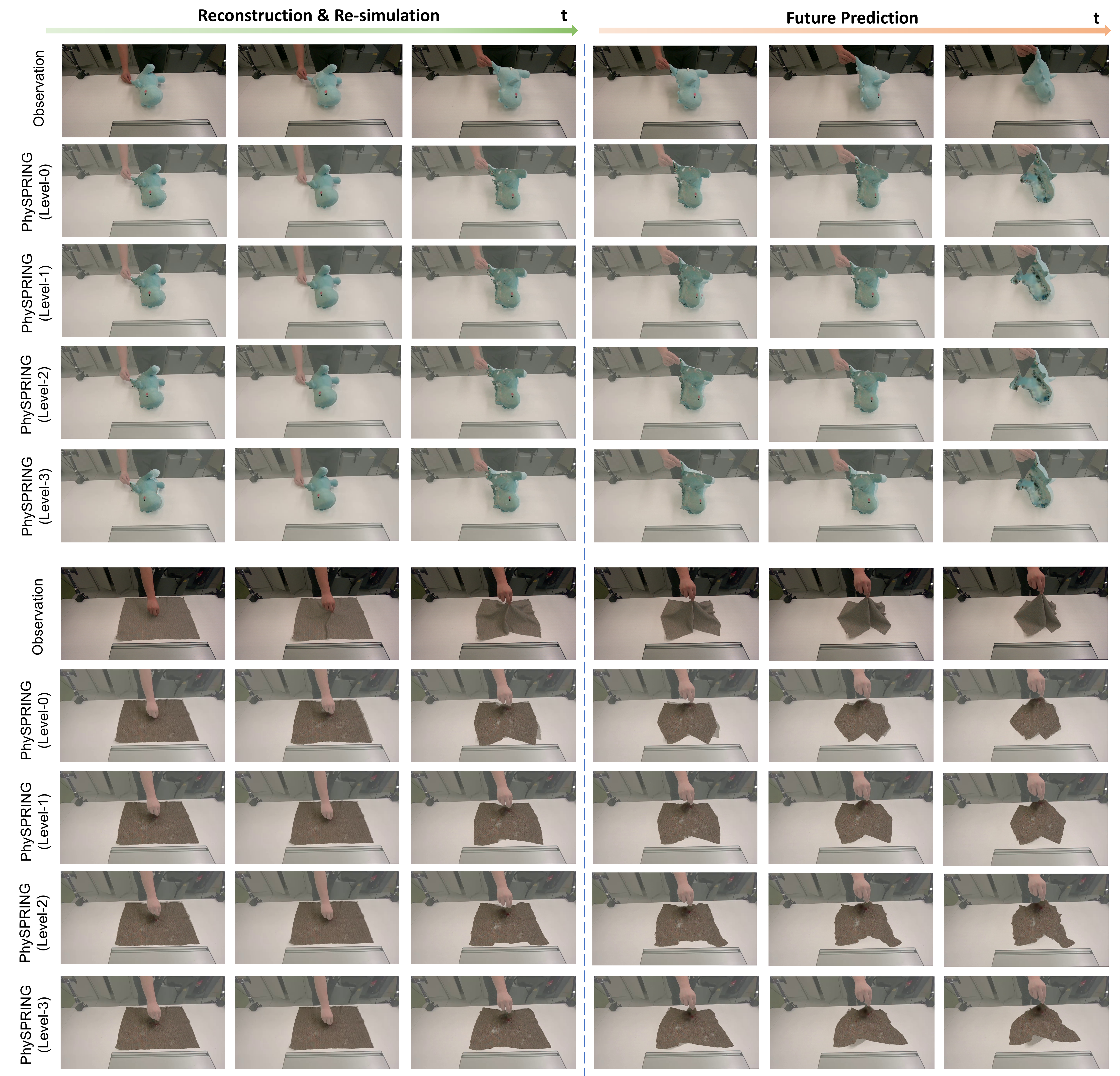}
\caption{Dinosaur and cloth visualizations across ROM levels.}
\label{fig:app_vis_roms}
\end{figure}

Fig.~\ref{fig:app_vis_roms} visualizes reconstruction\&re-simulation and future prediction for the dinosaur and cloth sequences across ROMs. Across Levels 0--3, \method{} preserves the main deformation patterns and temporal evolution of both objects with progressively reduced complexity. Although higher ROM levels introduce mild loss of fine geometric detail, the overall shape, motion trend, and visual consistency remain stable, demonstrating that the learned ROMs retain the essential dynamics needed for both re-simulation and prediction.

\subsection{Real2Sim evaluation}

\paragraph{Per-policy success counts across ROMs.}
\label{app:real2sim_success}

Table~\ref{tab:real2sim_success_rope} reports the current Real2Sim success-count sweep for the rope-routing task. We evaluate two policies, ACT~\citep{zhao2023learning} and $\pi_0$~\citep{black2024pi_0}, at different policy-training iterations, and compare the Real2Sim baseline simulator against the dense \method{} twin and its three reduced levels.

\begin{table*}[t]
\centering
\caption{Real2Sim rope-routing policy-evaluation success counts across ROM levels. Each entry reports the number of successful trials, and all entries use \textbf{27} tests.}
\label{tab:real2sim_success_rope}
\footnotesize
\setlength{\tabcolsep}{3.5pt}
\begin{tabular}{lc|ccccc}
\toprule
Policy & Iter. & Real2Sim & \makecell{Dense\\(L0)} & \makecell{Level 1} & \makecell{Level 2} & \makecell{Level 3} \\
\midrule
ACT & 300  & 6  & 4  & 5  & 3  & 6 \\
ACT & 500  & 7  & 6  & 2  & 4  & 3 \\
ACT & 700  & 18 & 17 & 9  & 12 & 9 \\
ACT & 1000 & 20 & 16 & 15 & 15 & 13 \\
ACT & 4000 & 19 & 24 & 22 & 20 & 21 \\
ACT & 7000 & 18 & 23 & 22 & 23 & 23 \\
\midrule
$\pi_0$ & 10000 & 14 & 12 & 14 & 14 & 12 \\
$\pi_0$ & 20000 & 21 & 21 & 25 & 24 & 20 \\
$\pi_0$ & 30000 & 26 & 24 & 24 & 24 & 24 \\
\bottomrule
\end{tabular}
\end{table*}

\paragraph{Action sampling effectiveness across ROMs.}
\label{app:real2sim_timing}

Table~\ref{tab:real2sim_action_speed} reports the action-sampling throughput measured in the Real2Sim loop.
For each reduced level $l$, we compute the relative throughput gain from FPS ratios:
\begin{equation}
S^{\mathrm{dense}}_l = \frac{\mathrm{FPS}_l}{\mathrm{FPS}_{\mathrm{dense}}}, \qquad
S^{\mathrm{r2s}}_l = \frac{\mathrm{FPS}_l}{\mathrm{FPS}_{\mathrm{Real2Sim}}}.
\end{equation}
Across both policies, progressively reducing the twin improves action-sampling throughput while preserving the policy-success counts reported in Table~\ref{tab:real2sim_success_rope}.

\begin{table*}[t]
\centering
\caption{Action-sampling throughput in the Real2Sim rope-routing loop. FPS is measured over successful evaluation cases; higher is better. Speed-ups are computed only for the reduced \method{} levels using the FPS-ratio formula above.}
\label{tab:real2sim_action_speed}
\footnotesize
\setlength{\tabcolsep}{6pt}
\renewcommand{\arraystretch}{1.08}
\begin{tabular}{llcccc}
\toprule
Policy & Simulator / twin & Nodes & FPS & \makecell{Speed-up\\vs Dense} & \makecell{Speed-up\\vs Real2Sim} \\
\midrule
\multicolumn{6}{l}{ACT} \\
& Real2Sim & 2002 & 4.872 & -- & -- \\
& Dense (L0) & 2002 & 4.597 & -- & -- \\
& Level 1 & 1243 & 5.159 & 1.12$\times$ & 1.06$\times$ \\
& Level 2 & 1050 & 5.560 & 1.21$\times$ & 1.14$\times$ \\
& Level 3 & 838  & 5.680 & 1.24$\times$ & 1.17$\times$ \\
\midrule
\multicolumn{6}{l}{\textbf{$\pi_0$}} \\
& Real2Sim & 2002 & 4.602 & -- & -- \\
& Dense (L0) & 2002 & 4.484 & -- & -- \\
& Level 1 & 1243 & 4.875 & 1.09$\times$ & 1.06$\times$ \\
& Level 2 & 1050 & 5.440 & 1.21$\times$ & 1.18$\times$ \\
& Level 3 & 838  & 5.660 & 1.26$\times$ & 1.23$\times$ \\
\bottomrule
\end{tabular}
\end{table*}

\paragraph{Motion-sequence visualization across ROMs.}
\label{app:real2sim_motion}

\begin{figure}[t]
\centering
\includegraphics[width=\textwidth]{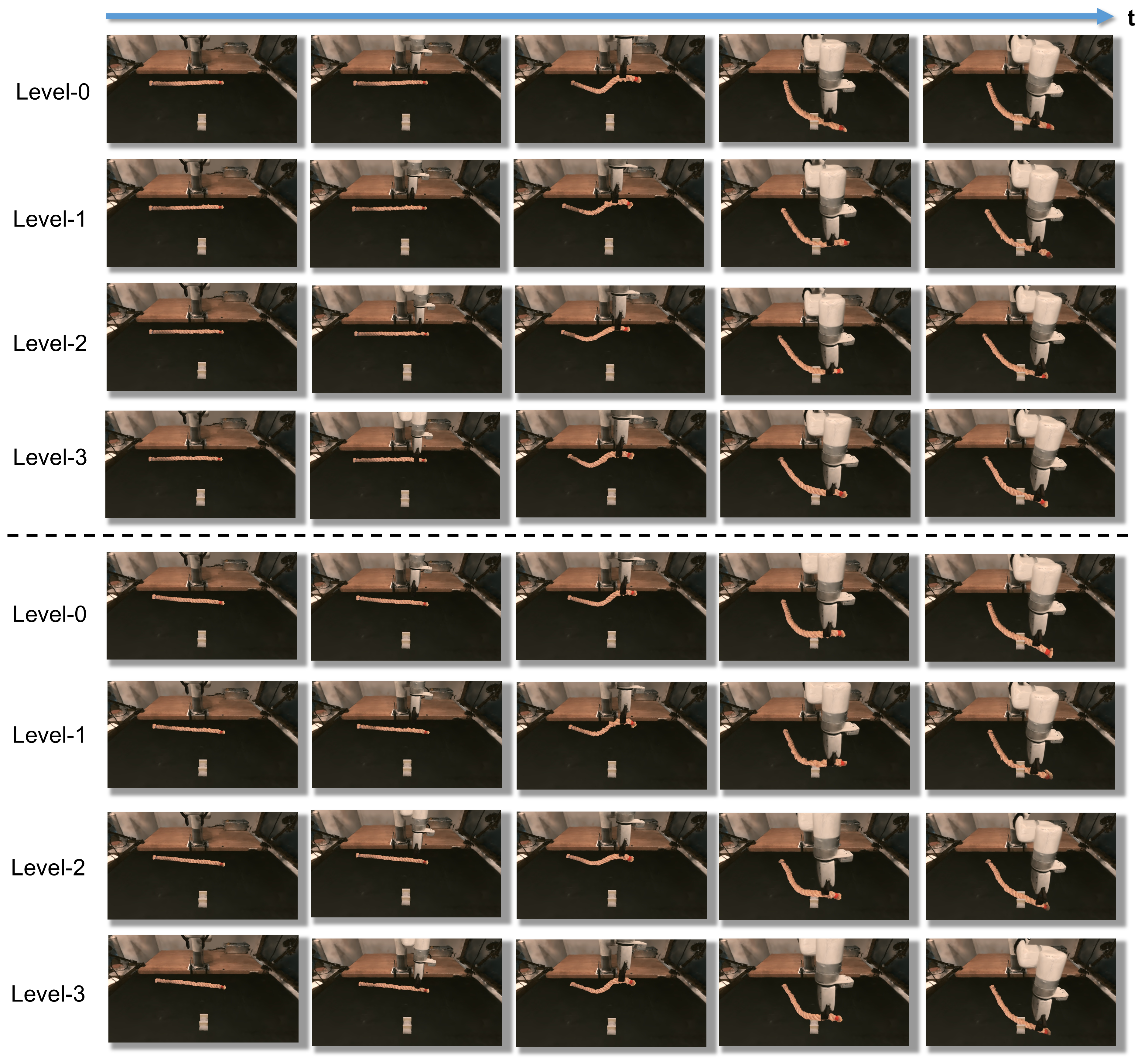}
\caption{Real2Sim rope manipulation visualizations across ROM levels. The two blocks show representative successful ACT motion sequences.}
\label{fig:app_real2sim_rope_levels}
\end{figure}

Fig.~\ref{fig:app_real2sim_rope_levels} visualizes two representative successful ACT demonstrations for the rope task in the Real2Sim~\citep{2025arXiv251104665Z} environment.
For each demonstration, rows compare the full-order twin (Level~0) with the progressively reduced twins (Levels~1--3), while columns show the temporal manipulation sequence.
Across all ROM levels, the rope motion, contact sequence, and final insertion behavior remain visually consistent, providing qualitative evidence that the reduced twins preserve policy-relevant deformation dynamics.

\end{document}